\documentclass[11pt,a4paper]{article}
\usepackage{times,latexsym}
\usepackage{url}
\usepackage[T1]{fontenc}

   \usepackage[acceptedWithA]{tacl2021v1}
\usepackage{tacl2021v1}

\usepackage{xspace,mfirstuc,tabulary}
\usepackage{graphicx}
\usepackage{boldline}
\usepackage{caption}
\usepackage{subcaption}
\usepackage{amsmath}

\newif\iftaclinstructions
\taclinstructionsfalse 
\iftaclinstructions
\renewcommand{\confidential}{}
\renewcommand{\anonsubtext}{(No author info supplied here, for consistency with
TACL-submission anonymization requirements)}
\newcommand{\instr}
\fi

\iftaclpubformat 

\else

\fi

\title{Dreams Are More ``Predictable'' Than You Think}

\author{
  Lorenzo Bertolini
  \\
  Department of Informatics, University of Sussex, Brighton BN1 9RH, UK
  \\
  \texttt{l.bertolini@sussex.ac.uk}
}

\date{}

\begin{document}
\maketitle




\begin{abstract}
A consistent body of evidence suggests that dream reports significantly vary from other types of textual transcripts with respect to semantic content. Furthermore, it appears to be a widespread belief in the dream/sleep research community that dream reports constitute rather ``unique'' strings of text. This might be a notable issue for the growing amount of approaches using natural language processing (NLP) tools to automatically analyse dream reports, as they largely rely on neural models trained on non-dream corpora scraped from the web. In this work, I will adopt state-of-the-art (SotA) large language models (LLMs), to study if and how dream reports deviate from other human-generated text strings, such as Wikipedia. Results show that, taken as a whole, DreamBank does not deviate from Wikipedia. Moreover, on average, single dream reports are significantly more predictable than Wikipedia articles. Preliminary evidence suggests that word count, gender, and visual impairment can significantly shape how predictable a dream report can appear to the model.
\end{abstract}




\section{Introduction}
Dream reports describe the content of our conscious experiences while asleep. Given their connection with our awakened state \cite{Blagrove2004} and conscious experience \cite{Nir2010, Siclari2017}, dream reports have long been of great interest to researchers and practitioners, and a variety of frameworks were built to study, analyse and annotate their content \cite{Hall-1966-content, hauri1975categorization, Schredl_10}. 

The analysis and annotation process of dream reports can be extremely time-consuming and relies upon human experts that usually undergo long training. This has limited the growth and reproducibility of research around dreams and dream reports \cite{Elce2021}. For this reason, researchers and practitioners have shown great interest in adopting natural language processing (NLP) tools to automatically analyse dream reports' content and structure (see \citet{Elce2021} for a review). Many of these approaches use machine-learning solutions trained on large amounts of text from the internet, like Wikipedia \cite{nadeau-etal-2006-automatic, Razavi2013, Altszyler2017, Sanz2018, McNamara-etal-2019-Dream, bertolini-etal-2023-automatic}.

While the extent to which dream reports differ from other forms of textual transcripts remains a matter of significant debate \cite{Kahan2011, domhoff2017emergence, Zheng2023}, evidence suggests that their semantic content and word use can significantly diverge from transcripts describing or concerning awakening (see \citet{Altszyler2017, Bulkeley2018, Zheng2023}, inter alia). However, it seems largely accepted by the research community that dream reports actually constitute a set of rather unique strings of text. Should this be the case, the growing practice of adopting NLP models trained on non-dream text could be significantly negatively impacted, especially when adopting unsupervised approaches, as already partially shown by \citet{bertolini-etal-2023-automatic}.

This work proposes to investigate this question using a fully unsupervised solution based on GPT-2 \cite{radford2019language}, a pre-trained autoregressive large language model (LLM), shown to efficiently model many linguistic structures and properties. The main question under investigation is related to the one proposed by \citet{Zheng2023}. However, instead of qualitatively identifying \emph{what content} makes a (limited set of) dream and waking reports different, I  study, in a quantitative manner, \emph{how much} a (large) set of dream reports seems ``surprising'' to a model that has seen a huge amount of non-dream-based text. This work makes four main contributions. First, it shows that, as a whole corpus, (a large proportion of) DreamBank is as predictable as (a comparable section of) Wikipedia. Second, at the single-text level, dream reports are on average significantly more predictable than Wikipedia articles. Third, it identifies a negative correlation between the number of words in a report/article and how ``surprising'' such a report/article appears to the model. Fourth, it provides preliminary evidence suggesting gender and visual impairment can significantly impact how ``surprising'' a report appears to the model. 




\section{Related Work} 
\citet{Elce2021} provide an exhaustive summary of work adopting computational approaches to the investigation of dream reports. As highlighted in the work, many of these approaches are based on dictionary-based frequency analysis of content words (e.g. \cite{Bulkeley2018, Mallett2021, Zheng2021, Yu2022, Zheng2023}). Of note, \citet{Bulkeley2018} propose a critical analysis of the use of a dictionary-based approach such as LIWC \cite{pennebaker-LIWC}, noticing, similarly to \citet{Zheng2023} how such a solution could be limited by problematic aspects of dream reports, such as typographical errors. However, the work strengthened the evidence that a method like LIWC could be used to discover differences between different types of dreams, such as nightmares, lucid dreams, and baseline dream reports. 

\citet{Zheng2023} expanded previous work by studying the differences between dream reports and other types of textual transcripts, using both LIWC and support vector machines \cite{cortes1995support} (SVM). The LIWC approach found a large set of categories that significantly differ between dream and non-dream reports, and the proposed SVM approach could successfully discriminate between the two categories of reports. However, the adopted dataset was quite limited in magnitude --- circa 800 instances, balanced between dream and non-dream reports. This limits the generalisability of the findings, largely grounding the observed difference to the dataset of choice. Moreover, since LIWC features were used as training and test data to the SVMs, it is not surprising that the model was able to discriminate between the two categories. 

Lastly, \citet{Altszyler2017} compared two word-embedding approaches (namely LSA \cite{Landauer1997AST} and word2vec's skip-gram with negative samples \cite{mikolov-etal-2013-efficent}) to investigate how the relationship between a content word like \emph{run} changes in large web corpora compared to a large collection of dream reports from DreamBank \cite{Domhoff2008}, and found that LSA better encoded the difference in the type of contexts such words appear in the two types of corpora. 

Overall, while previous work reveals qualitative information on \textit{what} makes a (frequently limited) set of reports different from a (relatively limited) set of textual transcripts, they do not reveal \textit{if} or \textit{how strongly} dream reports (quantitatively) deviate, on average, from other types of human-generated date. This work focuses on this last question and provides a first set of analyses on which macro-factors might impact how well dream reports can be modelled by current NLP tools.




\section{Experimental Setup}
\subsection{Model and measure}
The primary interest of this work is to investigate if dream reports are in fact harder to model and predict for a model like GPT-2. To study this question, I  adopt perplexity, which can be intuitively seen as a measure of how uncertain a model is about a given string of tokens \cite{chip2019evaluation}. While many other solutions have been proposed to evaluate how well a language model can capture different linguistic phenomena, perplexity is still widely used, and can surely inform us on how well a model reflects natural language \cite{meister-cotterell-2021-language}, and hence reveal how distant another string is to a more ``natural'' sequence. Since the goal is to understand if a machine trained on a very large amount of textual data ``perceives'' dream reports as ``surprising'' (i.e., as having a high perplexity), I  adopt GPT-2 (small) as the language model, using the Hugging Face \cite{wolf-etal-2020-transformers} implementation. When testing whole datasets, I  follow Hugging Face's source code, I  adopt a stride of 512, and a maximal input of 1024 tokens, as it produces the closest results to the original GPT-2 work\footnote{\url{https://huggingface.co/docs/transformers/perplexity}}. Of note, a very similar set-up was also adopted by \citet{Colla2022}, which showed how perplexity scores from GPT-2 and n-\emph{gram} models can be used to discriminate between healthy participants and patients with Alzheimer's disease.  

\subsection{Dream dataset}
Dream reports are extracted from a section\footnote{Available at \url{https://huggingface.co/datasets/DReAMy-lib/DreamBank-dreams-en}} of DreamBank \cite{Domhoff2008}. The final dataset contains approximately 22k reports in the English language, annotated with respect to gender, year of collection, and Series --- subsets of DreamBank representing (groups of) individuals from which dreams are collected. 

\subsection{Text baseline}
\begin{figure}[htb!]
    \centering
    \includegraphics[width=\columnwidth]{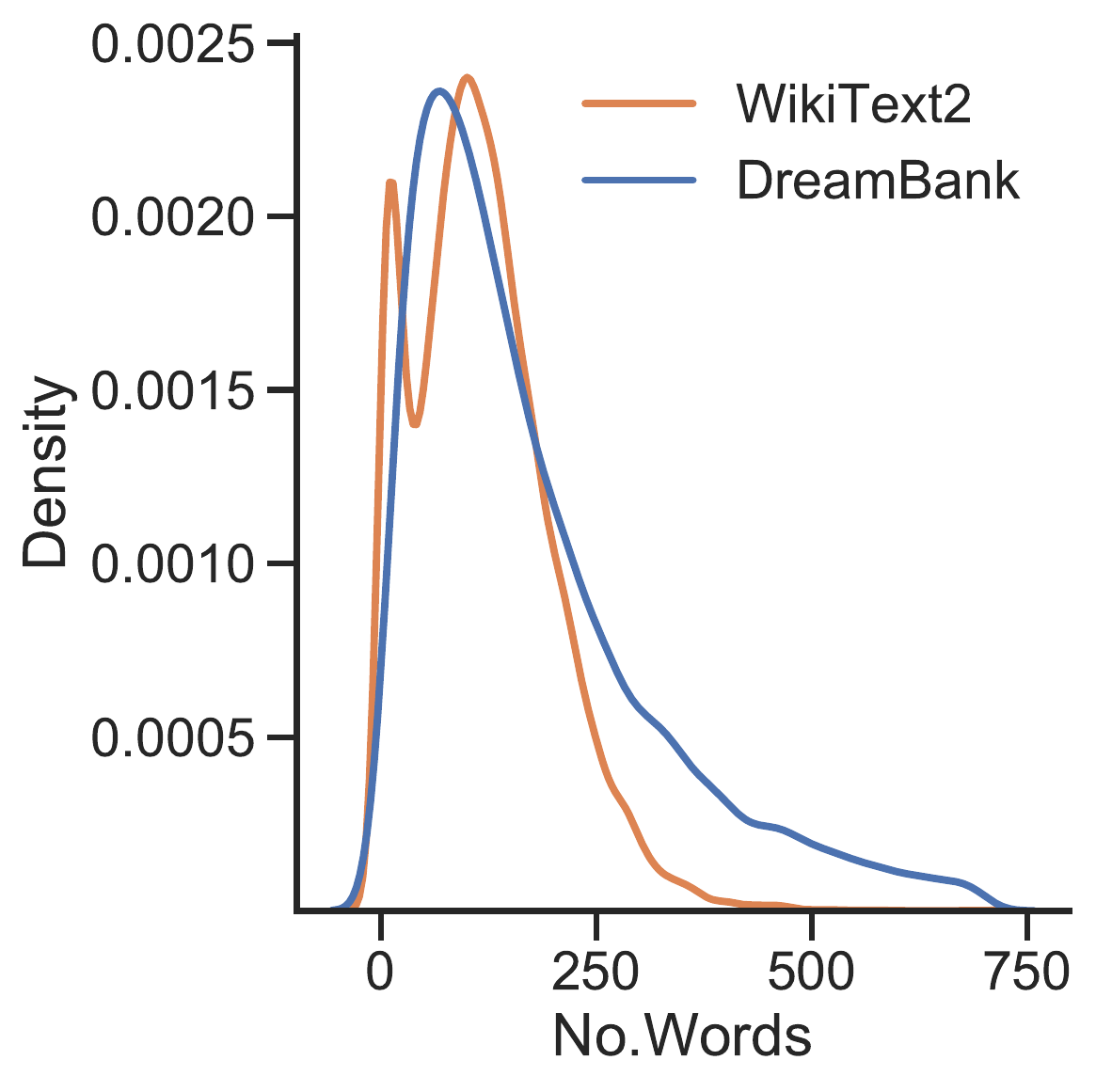}
    \caption{\label{fig:no_words_per_report} Density distributions of word counts (No.Words, x-axis) per textual report/article in DreamBank and WikiText2 (train).}
\end{figure}

As a baseline, I  consider the perplexity produced by the model on the WikiText2 dataset \cite{merity2017pointer}, from which the article's titles were removed. The choice is motivated by three reasons. First, Wikipedia was not included in the data used to train the selected model, limiting the possibility of data contamination. Second, it allows for a comparison with a more standardised text in terms of syntactic and semantic structure. Large parts of Wikipedia are formally structured and heavily curated, and can hence work as a ``stress'' test for the hypothesis that dream reports are notably different. Third, while statistically different ($p$ < .001), the distributions of the number of words per text (No.Words) are not substantially different, as shown in Figure \ref{fig:no_words_per_report}.

\subsection{Analyses}
Throughout the work, distributions are compared with a random permutation test. Reported $p$ values refer to the corrected score, obtained via Holm \cite{Holm1979ASS} correction. All experiments are run with the support of an RTX 3090 GPU. Code and collected data are freely available here\footnote{\url{https://github.com/lorenzoscottb/dream_perplexity}}. 




\section{DreamBank vs. WikiText2}
\begin{figure}[htb!]
    \centering
    \includegraphics[width=\columnwidth]{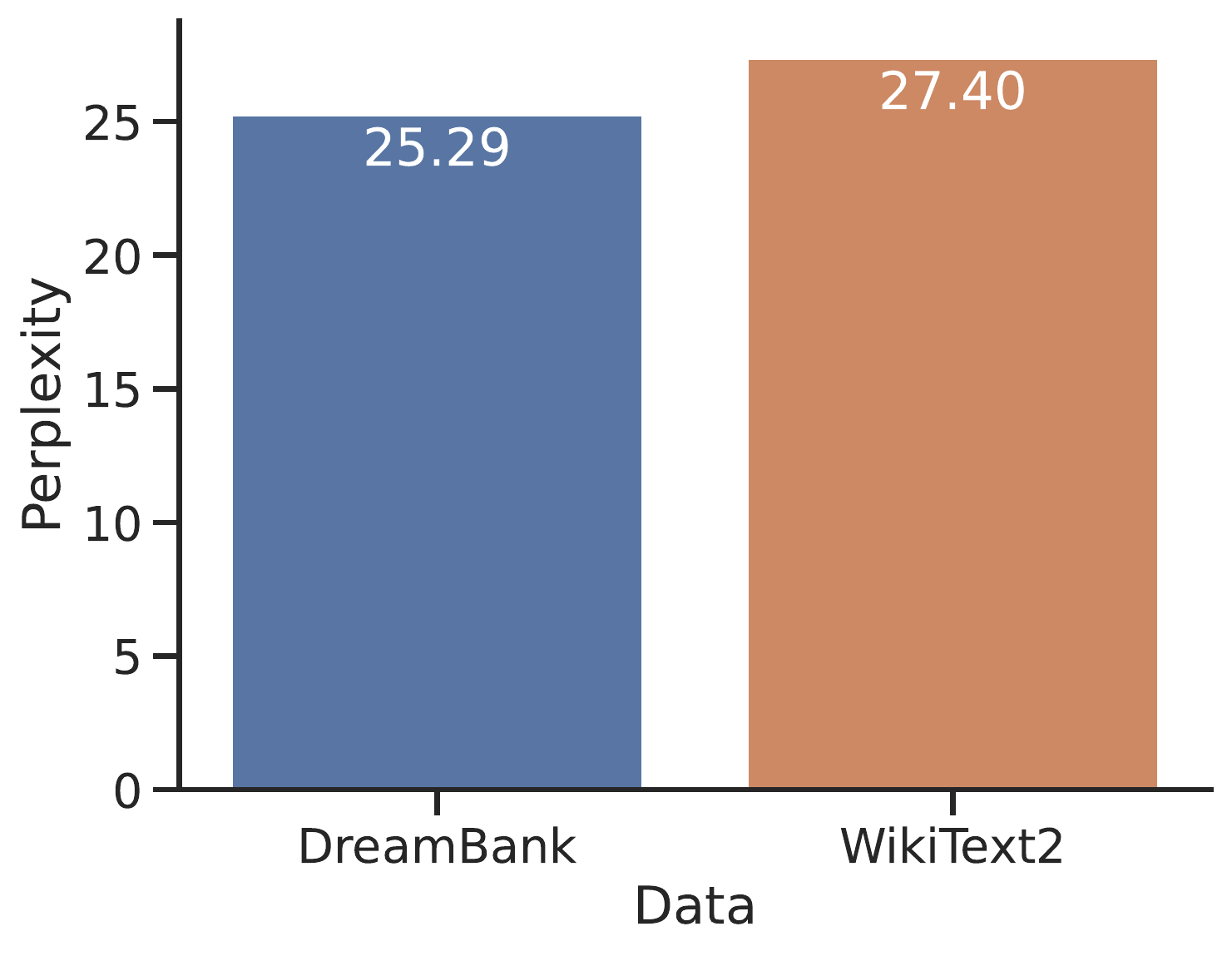}
    \caption{\label{fig:db_wiki_bar} Overall perplexities. Perplexities scores (lower is better) obtained by GPT-2 on the whole DreamBank and (train) WikiText2 datasets.}
\end{figure}

The first experiment focuses on the overall perplexity of DreamBank and WikiText. That is, the two datasets of choice are compared in their entirety. For a fair comparison, DreamBank ($\sim$22k items) is tested against the train split of (title-less) WikiText2 ($\sim$18k items). The maximum input sequence to the model is here set to 1024 tokens, with a stride of 512 tokens. Figure \ref{fig:db_wiki_bar} reports the perplexities produced by the model on the two datasets and shows how the two datasets, comparable in size and tokens-per-text, also obtain comparable results in terms of general perplexity. In other words, considered as a whole, DreamBank appears as predictable as (a subset of) Wikipedia to GPT-2.

\begin{figure*}[tbhp]
\centering
\includegraphics[width=\textwidth]{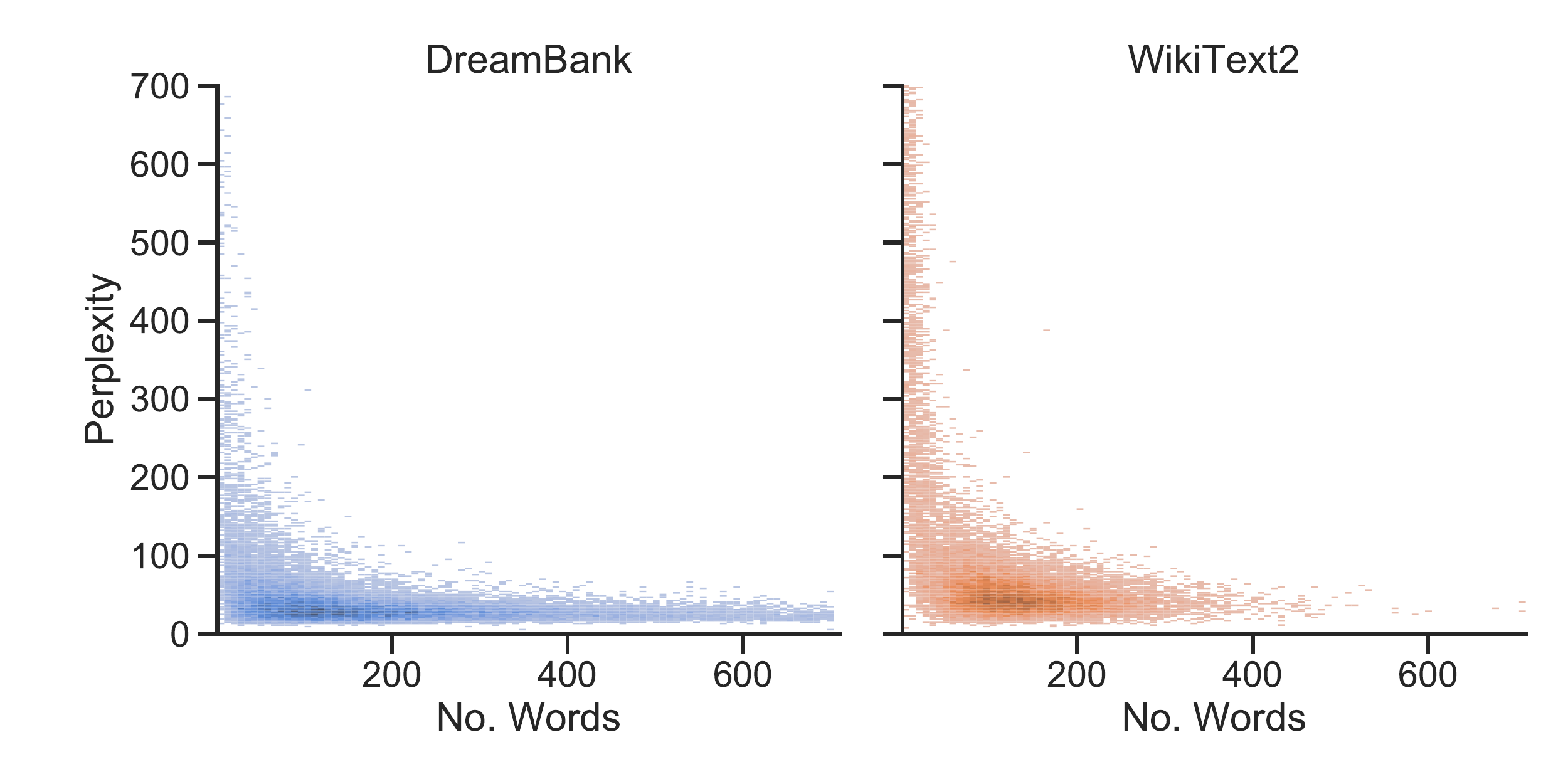}
\caption{Perplexity and the number of words (No. Words) correlation analysis. Visualisation of the interaction between the number of words in a given text (x-axis) and the perplexity score (y-axis) generated by GPT-2 for each text, divided by datasets (columns, DreamBank vs. Wikitext2 (train).}
\label{fig:nowds_ppl_displot}
\end{figure*}

\begin{figure}[htb!]
    \centering
    \includegraphics[width=\columnwidth]{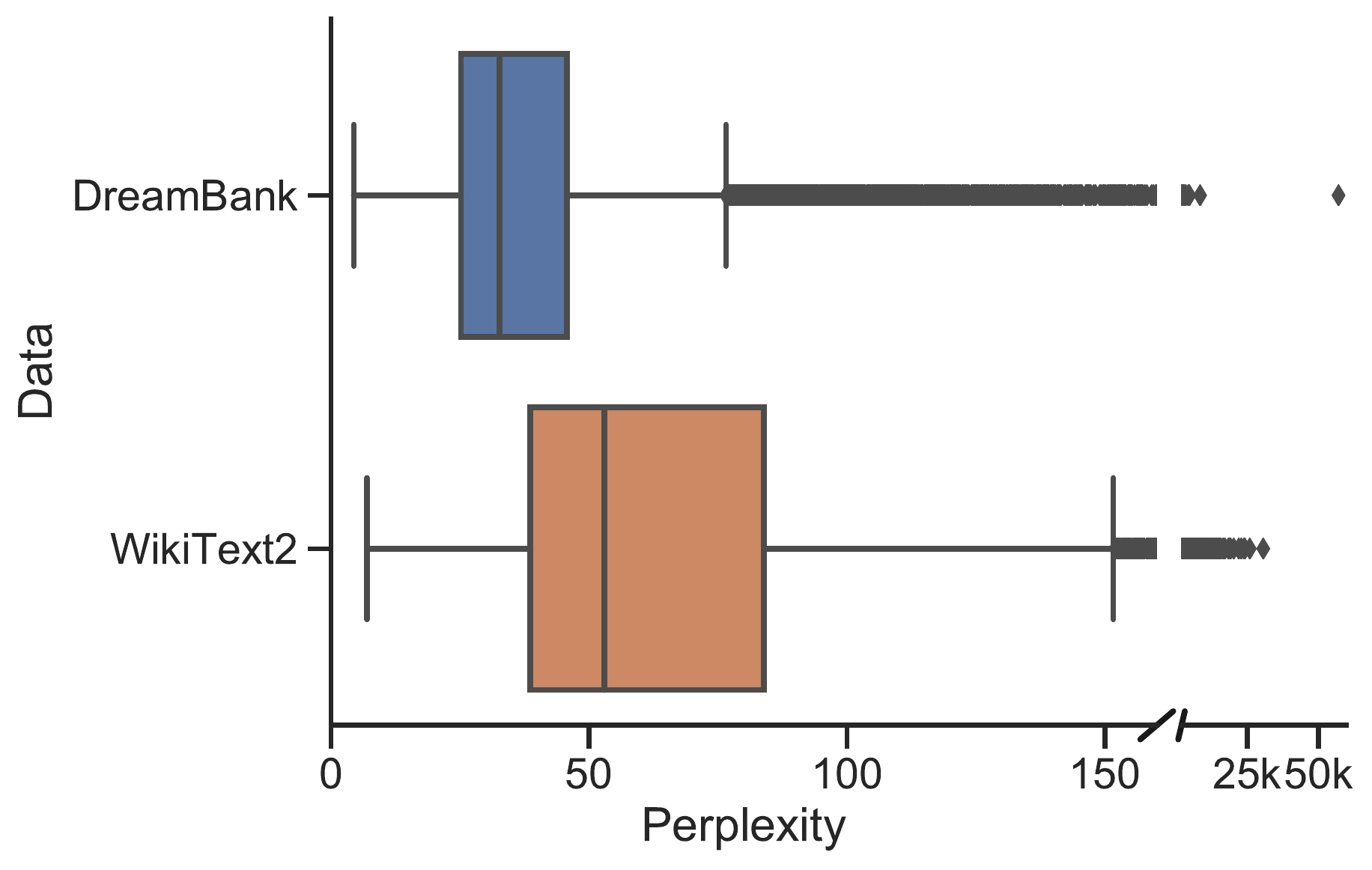}
    \caption{\label{fig:wiki_db_reports_ppl}Per-report perplexities. Distributions of perplexities scores obtained by GPT-2 on DreamBank and (train) WikiText2 on different reports separately.}
\end{figure}

Figure \ref{fig:wiki_db_reports_ppl} reports the distributions of the perplexity scores generated by GPT-2 for each dream report and WikiText2 (train) articles. The distributions are significantly different ($p$ < .001), and show how, overall, dream reports present a lower and less varied perplexity, compared to articles from Wikipedia.




\section{Impact of Macro-Factors}
This section concerns the impact of the immediately available features of dream reports (i.e., number of words, gender and series, year of collection, and visual impairment) on the perplexity. Since series contain reports from participants that identify themselves as the same gender, the investigation of gender and series is presented together. For the same reason, and given the imbalance of visual impairment (i.e., blind vs. normal-sighted participants) and gender distributions, the statistical analyses are performed separately, instead of using a general logistic or linear model.

\begin{figure}[htb!]
    \centering
    \includegraphics[width=\columnwidth]{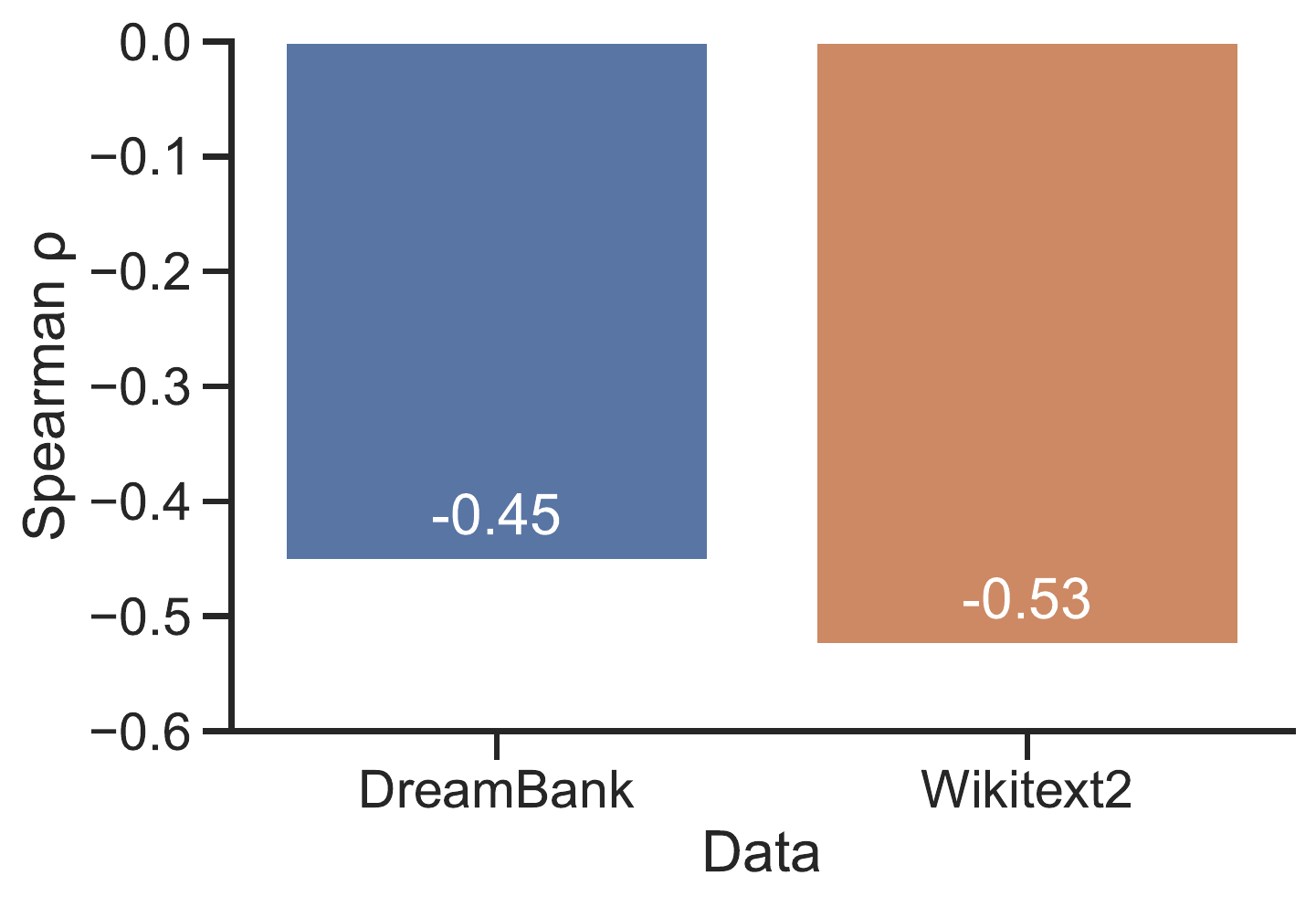}
    \caption{\label{fig:nowds_ppl_corr} Correlations (in Spearman's $\rho$) between perplexity and word count (No.Words), divided by datasets.}
\end{figure}

\begin{figure*}[tbhp]
\centering
\includegraphics[width=\linewidth]{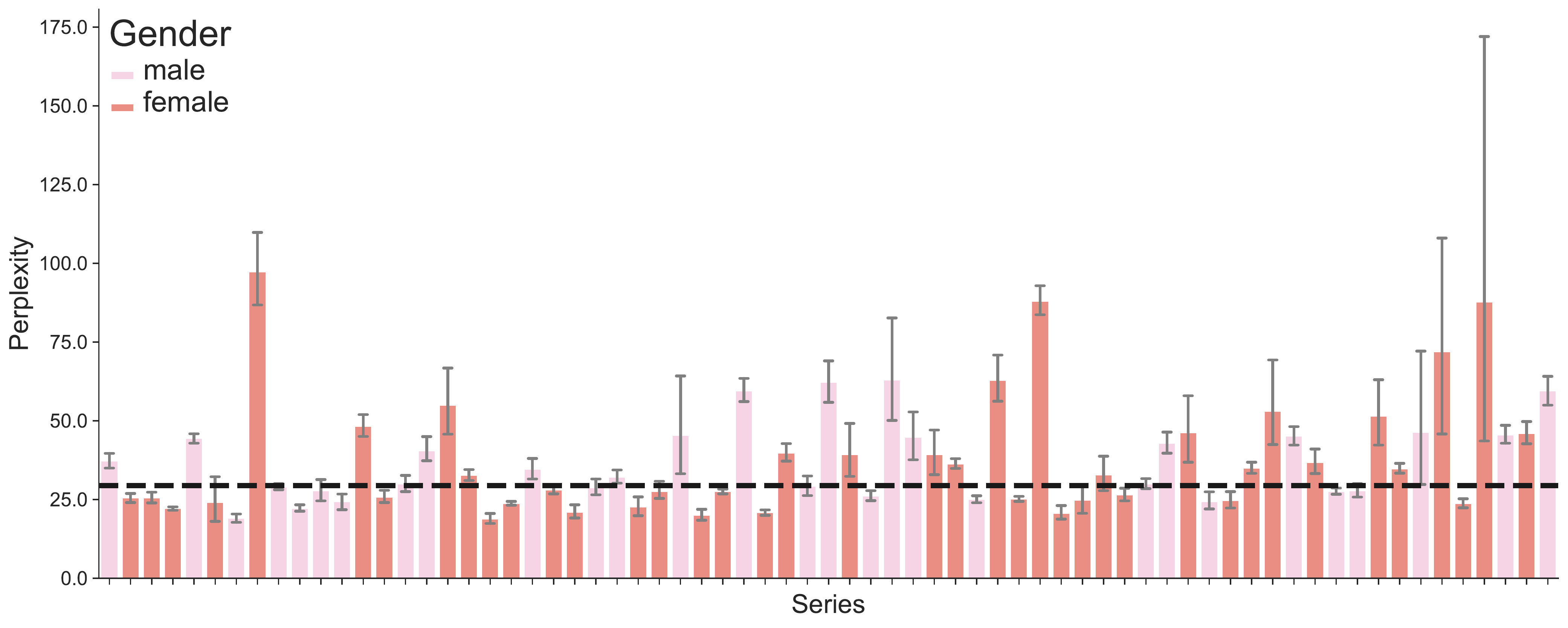}
\caption{Perplexity by series (and gender). Average (± standard deviation) of DreamBank's perplexity score extracted from GPT 2, divided by Series (x axis) and gender (hue). The dashed line reports the perplexity of the whole WikiText2 (test) dataset.}
\label{fig:general_perplexoty}
\end{figure*}

\subsection{Word count}
Figure \ref{fig:nowds_ppl_corr} reports the results of the correlation analysis between the number of words per report (No.Words) and perplexity scores. For comparison, the same analysis is conducted on WikiText2 (train) articles. As shown in the figure, both (significant, $p$ < .0001) correlations are markedly negative. In other words, as reports/articles grow in the number of words, the perplexity scores generated by GPT-2 decrease. Intuitively, this suggests that longer reports are easier to predict for GPT-2. However, as shown by the distribution plot in Figure \ref{fig:nowds_ppl_displot}, the cores of the distributions are notably close to the origin, and most of the distribution seems parallel to the x axis (No.Words), especially in the case of DreamBank. This suggests that the effect might be strongly modulated by outliers that are impacted by other factors than the number of words constructing a report/article.

\subsection{Gender and series}
\begin{figure}[htb!]
    \centering
    \includegraphics[width=\columnwidth]{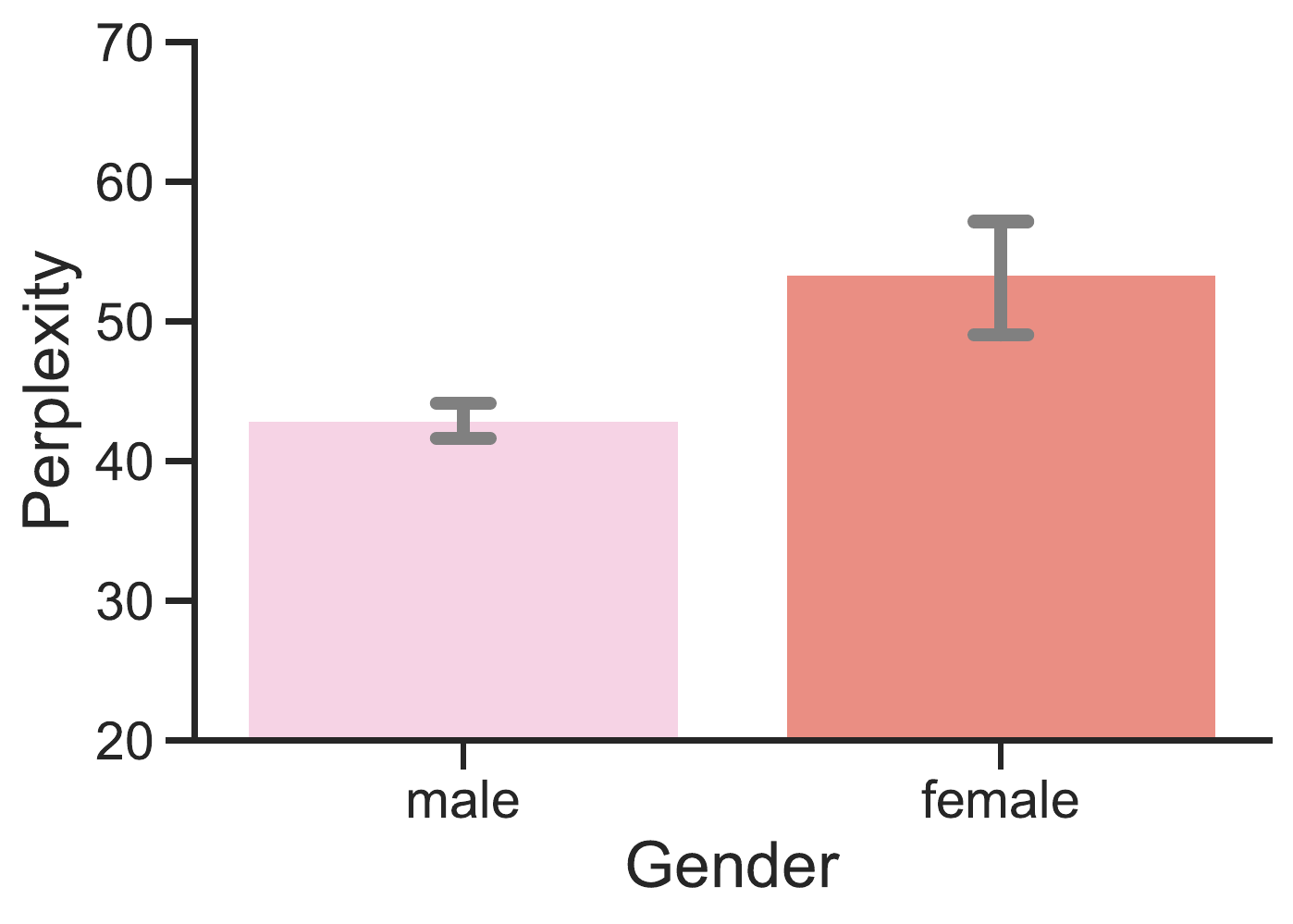}
    \caption{\label{fig:db_ppl_gender}Per-report perplexities: gender. Distributions of perplexities scores obtained by GPT-2 on DreamBank single dream reports divided by the gender of the participants.}
\end{figure}

\begin{figure}[htb!]
    \centering
    \includegraphics[width=\columnwidth]{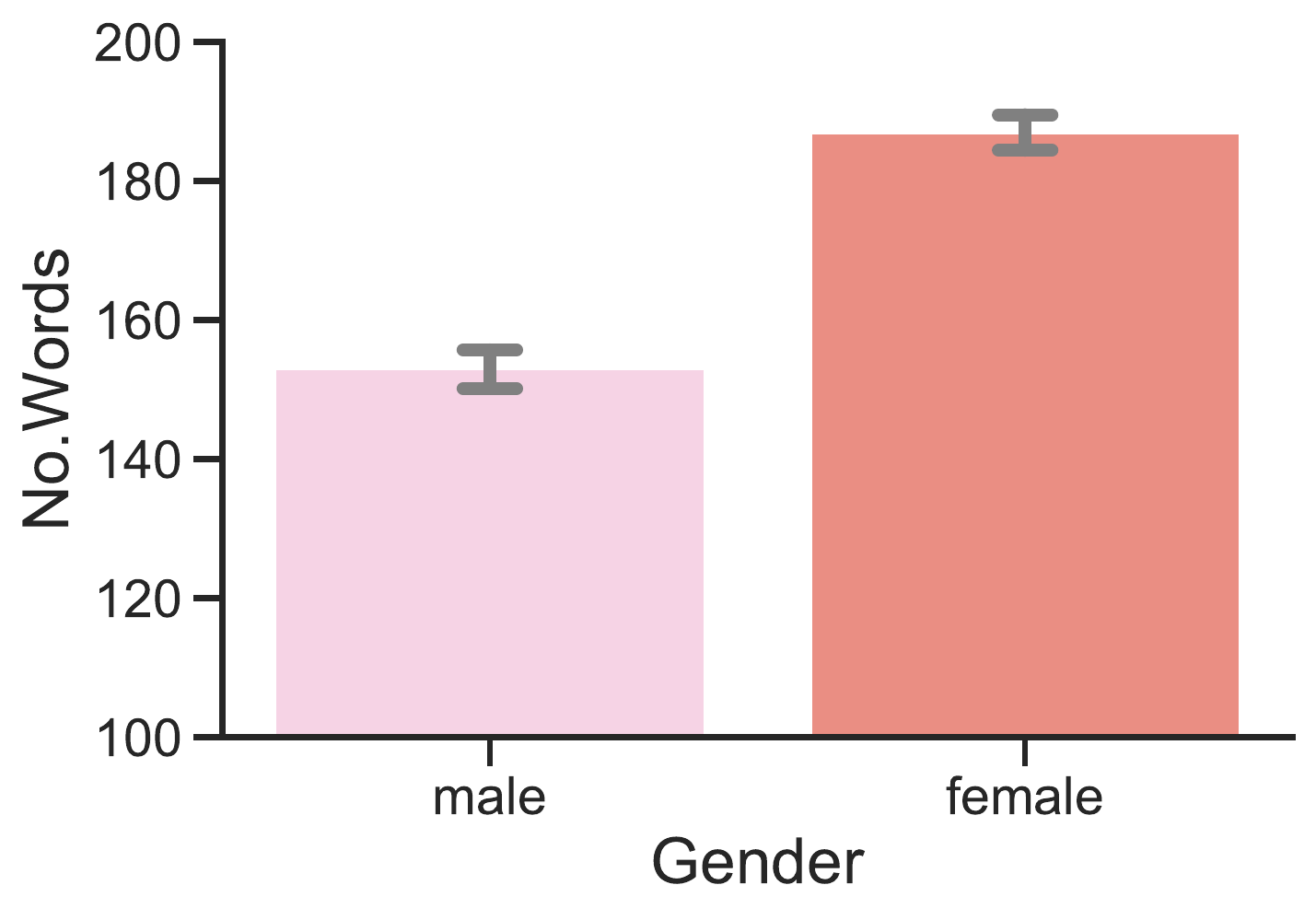}
    \caption{\label{fig:db_nw_gender}Per-report number of words: gender. Distributions of the word count of DreamBank single dream reports divided by the gender of the participants.}
\end{figure}

Figure \ref{fig:db_ppl_gender} compares perplexity scores obtained for reports from male and female participants. Male-generated reports are significantly ($p$ < .05) easier to predict than female ones and present less variance in the perplexity scores. Given the mentioned negative relation between perplexity scores and the number of words, one might expect to see male reports to be on average significantly longer than female ones. However, the observed trend is exactly the opposite, as shown in Figure \ref{fig:db_nw_gender}. That is, male participants write notably (and significantly, $p$ < .001) shorter reports than females, as already observed in other work (e.g., \citet{Mathes2013GenderDI})

A correlation analysis based on Spearman's $\rho$ did not find a relevant interaction between DreamBank's series and perplexity ($\rho$=–0.13, $p$ < .0001) or word count ($\rho$=–0.09, $p$ < .0001). However, as suggested by Figure \ref{fig:general_perplexoty}, it is possible that the higher and less stable perplexity observed in female participants is impacted by a small set of series, with a high average perplexity  and strong inner variance. Moreover, it is interesting to notice how almost half --- 49\% --- of all series actually present an average perplexity score comparable to the one of WikiText2 (test).

\subsection{Year of collection}
The effect of the year of data collection on the perplexity scores was also assessed with correlation analysis. To better analyse and interpret the relation between perplexity and the year of collection, the categorical framing of the variable (e.g., ``1980s - 1990s'') was converted into a continuous one, by simply finding the average year of collection (e.g, \texttt{1985}). Instances with non-available data were excluded from the analysis. 

The obtained value, together with the original ones, are presented in Table \ref{tab:year_conv} of Appendix \ref{app_y}. The result of the analysis suggests a negative correlation between the year of collection and perplexity scores. In other words, as one might expect, reports produced in more recent years are easier to model for GPT-2, and hence tend to produce lower perplexity scores. However, while strongly significant ($p$ < .0001), the effect was very weak ($\rho$=–.17).

\subsection{Vision impairment}
Lastly, I  consider a final variable that allows us to discriminate between dream reports: visual impairment. Among the series of DreamBanks, two of them collect reports from (male and female) blind participants. To compare this set with reports with one collected from sighted participants, I sampled a set of reports from DreamBank reports having the same range of perplexity scores observed for blind participants.

\begin{figure}[htb!]
    \centering
    \includegraphics[width=\columnwidth]{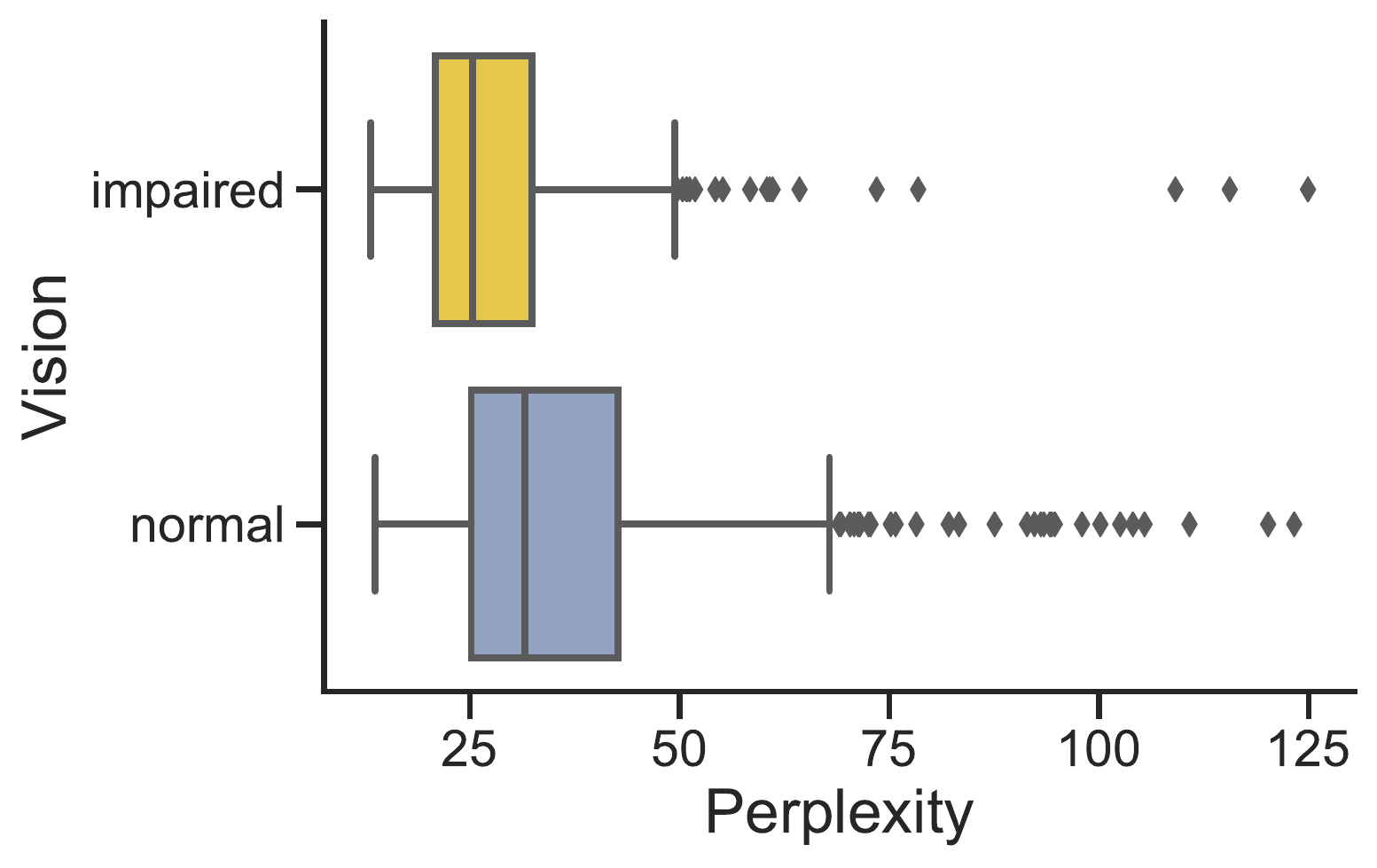}
    \caption{\label{fig:blinf_normal_box}Per-report perplexities: vision impairment. Distributions of perplexities scores obtained by GPT-2 on DreamBank single dream reports divided by vision impairment of the participants.} 
\end{figure}  

\begin{figure}[htb!]
    \centering
    \includegraphics[width=\columnwidth]{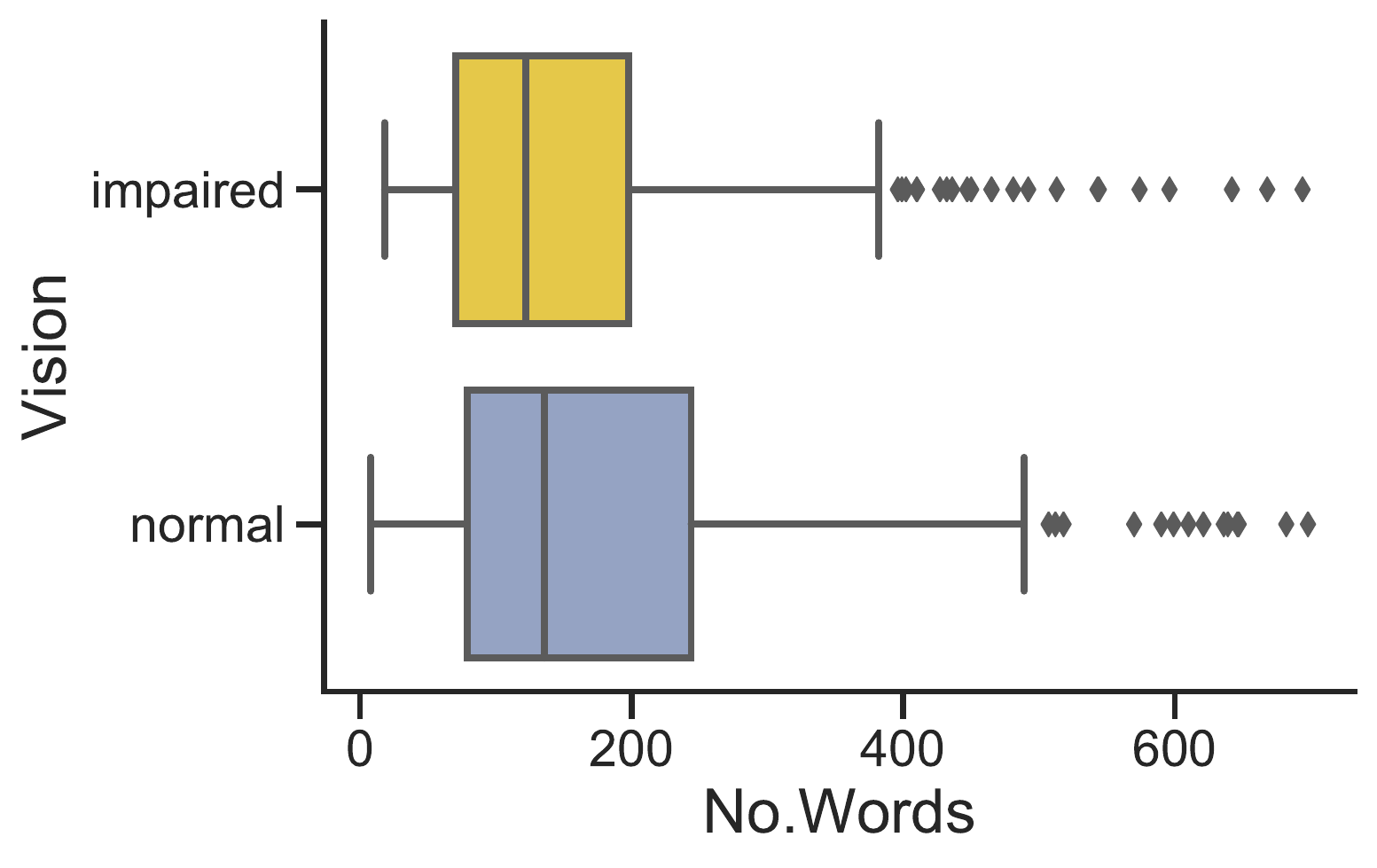}
    \caption{\label{fig:blinf_normal_box_nw}Per-report number of words: vision impairment. Distributions of the word count (No.Words) for DreamBank single dream reports divided by vision impairment of the participants.} 
\end{figure}

Figure \ref{fig:blinf_normal_box} presents the comparison of perplexity scores distributions produced by visually impaired and sighted participants. The two distributions significantly ($p$ < .001) and notably differ with respect to their perplexity scores. As shown in the figure, even when sampling from a limited range of perplexity scores, values for visually impaired participants are notably lower, have a notably smaller variance, and have a median score lower than WikiText2 (test) perplexity. The two classes of reports also significantly differ in terms of length ($p$ < .05). Similar to what was observed for gender, the class with the lower average perplexity --- visually-impaired participants --- is also the class producing (slightly) shorter reports, as shown in Figure \ref{fig:blinf_normal_box_nw}.

\section{Discussion}
A growing amount of work has adopted NLP tools to investigate and annotate dream reports (see \citet{Elce2021, bertolini-etal-2023-automatic} for more details). Many of these approaches rely on neural models of various dimensions, trained on large text corpora scraped from the web \cite{radford2019language}. While a consistent body of evidence has shown that the semantic content of dream reports significantly differs from other types of textual transcripts (see \citet{Altszyler2017, Bulkeley2018, Zheng2023}, inter alia), the dream/sleep research community seems to believe that they also constitute a rather ``unique'' set of textual strings. Should this be the case, dream reports might be difficult to model and predict for many NLP models, especially if used without supervision --- as already hinted by \citet{bertolini-etal-2023-automatic}.

In this work, I  adopted a state-of-the-art large language model (LLM) to test such a hypothesis. More specifically, I  studied how well GPT-2 (small), can model and predict dream reports, compared to a more ``standard'' piece of text like Wikipedia. Using perplexity as a measure of uncertainty, the work has shown that, taken as a whole, DreamBank is not harder to model than Wikipedia. Moreover, perplexity scores were found to be significantly lower (hence better) for single dream reports than for single Wikipedia. This suggests that GPT-2 pre-training, based on a web-scraped corpus, is likely enough to allow the model to manipulate dream reports without further supervision. Furthermore, it could be evidence of the fact that the findings from \citet{bertolini-etal-2023-automatic} might be more related to the specific sentiment-analysis dataset used for fine-tuning, rather than to the absence of direct supervision. 

The main findings of this work suggest that dream reports are not a unique and unpredictable class of textual strings per-se. However, just like Wikipedia's articles, some \textit{can be}. The work has hence proposed a first set of analyses to understand what might make a report more predictable. The focus was on the four variables immediately measurable from the adopted dataset: word count (No.Words), series and gender, year of collection, and visual impairments. The correlation analysis found a (rather expected) negative interaction between No.Words and perplexity scores. However, a more detailed investigation suggested that the observed effect might be largely influenced by a consistent set of outliers. In other words, there appears to be another mediating variable influencing how difficult it is for GPT-2 to model a dream report. Overall, the analysis further weakened the hypothesis that dream reports are rather unique strings of texts. All DreamBank's results, from the negative correlation to the outliers' effect and the shape of the distribution, found a strong match in the results produced by the model when tested on Wikipedia data.

The analysis of the results based on gender and visual impairment further challenged the strength of the negative correlation between perplexity and word count. In both cases, the group with significantly lower perplexity scores (i.e., male and blind participants) also tend to produce significantly shorter reports. This suggests that the real issue is less related to the actual number of words and more to the \textit{kind} of words. A similar conclusion was also proposed in \citet{bertolini-etal-2023-automatic}. Using an out-of-distribution ablation experiment, it was shown that leaving a specific DreamBank series out of training made it difficult for the model to handle a specific emotion (e.g., happiness for the \texttt{Bea 1} series.). The authors noted that this could not be simply explained by the number of instances in the training data, and was likely related to the specific vocabulary used in that series to describe that specific emotion.

The work also adds more evidence to the existing body of work investigating gender-based differences in dream reports \cite{Hall-1966-content, SCHREDL2008, Wong2016AutomaticGD}. While repeatedly observed, these differences were mainly constrained to a report's semantic content and/or grammatical structure, such as a reference to a specific emotion, use of violent language, or part of speech use. This work suggests that the observed distinction might have a very tangible effect since reports produced by male dreamers were found to be significantly easier, on average, to model by GPT-2. This likely suggests that the distinction is even deeper than previously noted, and might include a combination of content, vocabulary, and structure.

The result suggesting that blind dreamers produced more predictable reports seems more difficult to frame in the current literature and knowledge. Multiple pieces of evidence across time have shown how blind participants express a significantly lower amount of visual features in their reports, predominately presenting auditory, tactile and olfactory reference \cite{kirtley1975psychology, Hurovitz1999TheDO, Meaidi2014}. However, \citet{Meaidi2014} showed that these differences can significantly vary between congenitally and late blind, participants, and both series contain a mixture of congenitally and non-congenitally blind participants (although most have been for more then 20 years). A possibility might be that maintaining access to the visual modality while dreaming allows for a larger degree of abstraction and variance of dream content, leading sighted participants to generate more diverse reports, that can result in harder sequences to predict for the model.

Concerning the year of collection of each report, one might find the neglectable effect of this variable as unexpected, especially considering that many reports were collected at a time when the internet existed only in the minds of visionary scientists and writers. However, this might be easily explained by the fact that the internet is now a collection of extremely heterogeneous documents, that obviously include very old textual instances. It is hence possible that, while specific reports did not leak into the training data, their vocabulary and style might very well have. In other words, the model might have been exposed also to the form and vocabulary used in older reports.

To conclude, it is important to notice that this work has three main limitations. First of all, since WebText, the dataset used to train GPT-2, is not open-source, it is hard to estimate possible data leakage. That is, whether a part of the test data used in this work was also included in the training data for the model. In their work, \citet{radford2019language} note that training text was scraped following outbound links from
Reddit, with at least 3 karma, and one link connecting Reddit to DreamBank. However, the link reached the main page of DreamBank, which does not allow scraping dream reports. As shown by example codes (e.g., here\footnote{\url{https://github.com/mattbierner/DreamScrape}}), the main solution to acquire dream reports from  DreamBank is to iteratively sample them via the \texttt{random sample} page, which requires actively entering specific settings --- such as series or number of words --- to print out a set of reports. In other words, it seems quite unlikely that a consistent part of the test data for this work was in fact also included in the training data for GPT-2. Second, the language of tested items was limited to English. Third, the adopted dream report dataset, DreamBank, is not fully transparent about the extent to which the reports were manipulated. The extended amount of grammatical errors and unformal structures/forms found upon a manual inspection of a (limited) set of reports suggested that the data went through a very limited manipulation, but this can not be widely confirmed. Future work will have to investigate how strongly these findings can be generalised to other languages and dream reports datasets, as well as provide a more detailed explanation of what might make a report more complex to predict for a current LLM like GPT-2, taking more into consideration semantic content and syntactic structures. 




\section{Conclusion}
This work proposed adopting large language models (LLM) to test the hypothesis that dream reports are, a priori rather unique strings of texts, and are hence difficult to model and predict for current NLP tools trained on large non-dream corpora scraped from the web. Using perplexity as an uncertainty measure, and GPT-2 as a model, the work has shown that, taken as a whole, predicting and modelling DreamBank is not harder than doing the same with Wikipedia. Furthermore, when considered separately, dream reports are on average significantly easier to model compared to single Wikipedia articles. The study then presented a detailed analysis of how reports' length and other macro-factors of DreamBank's reports shape how easy it is for the selected model to predict a report.  




\section*{Acknowledgements} 
This research was supported by the EU Horizon 2020 project HumanE-AI (grant no. 952026). I would like to thank Adriana Michalak for her comments on a previous draft of the work.

\bibliography{tacl2021, anthology}
\bibliographystyle{acl_natbib}

\newpage




\appendix
\section{Year of Collection}\label{app_y}

\begin{table}[htb!]
\centering
\resizebox{.8\columnwidth}{!}{%
\begin{tabular}{lc}
\hlineB{3}
\textbf{DreamBank} & \textbf{Integer Conversion} \\
\hlineB{3}
1897-1918 & 1907 \\
1912-1965 & 1938 \\
1939 & 1939 \\
1940-1998 & 1969 \\
1940s-1950s & 1945 \\
1940s-1950s \& 1990s & 1960 \\
1946-1950 & 1948 \\
1948-1949 & 1948 \\
1949-1964 & 1956 \\
1949-1997 & 1973 \\
1957-1959 & 1958 \\
1960-1997 & 1978 \\
1960-1999 & 1979 \\
1962 & 1962 \\
1963-1965 & 1964 \\
1963-1967 & 1965 \\
1964 & 1964 \\
1968 & 1968 \\
1970 & 1970 \\
1970-2008 & 1989 \\
1971 & 1971 \\
1980-2002 & 1991 \\
1985-1997 & 1991 \\
1990-1999 & 1994 \\
1990s & 1990 \\
1991-1993 & 1992 \\
1992-1998 & 1995 \\
1992-1999 & 1995 \\
1995 & 1995 \\
1996 & 1996 \\
1996-1997 & 1996 \\
1996-1998 & 1997 \\
1997 & 1997 \\
1997-1999 & 1998 \\
1997-2000 & 1998 \\
1997-2001 & 1999 \\
1998 & 1998 \\
1998-2000 & 1999 \\
1999- & 2010 \\
1999-2000 & 1999 \\
1999-2001 & 2000 \\
2000 & 2000 \\
2000-2001 & 2000 \\
2001-2003 & 2002 \\
2003-2004 & 2003 \\
2003-2005 & 2004 \\
2003-2006 & 2004 \\
2004 & 2004 \\
2007-2010 & 2008 \\
2009 & 2009 \\
2010-2011 & 2010 \\
? & NaN \\
late 1990s & 1998 \\
mid-1980s & 1985 \\
mid-1990s & 1995 \\
\hlineB{3}
\end{tabular}
}
\caption{\label{tab:year_conv}Conversion table for DreamBank's year of collection variable.}
\end{table}

\end{document}